\documentclass{article}

\usepackage{microtype}
\usepackage{graphicx}
\usepackage{subfigure}
\usepackage{booktabs} 
\usepackage{xcolor}
\usepackage{hyperref}

\usepackage[skins]{tcolorbox}
\usepackage{caption}
\usepackage{tabularx}
\usepackage{float}
\usepackage{lipsum}

\usepackage{xcolor}

\hypersetup{
    colorlinks=true, 
    linkcolor=red,  
    filecolor=magenta, 
    citecolor=black,
    urlcolor=cyan,   
}

\usepackage[accepted]{icml2023}

\usepackage{amsmath}
\usepackage{amssymb}
\usepackage{mathtools}
\usepackage{amsthm}
\usepackage{multirow}
\usepackage{multicol}
\usepackage[capitalize,noabbrev]{cleveref}

\theoremstyle{plain}

\theoremstyle{definition}

\theoremstyle{remark}

\usepackage{pifont}       
\usepackage{bbding}       
\usepackage{fontawesome}
\usepackage[textsize=tiny]{todonotes}

\icmltitlerunning{Proximity QA: Unleashing the Power of Multi-Modal Large Language Models for Spatial Proximity Analysiss}

\begin{document}

\twocolumn[
\icmltitle{Proximity QA: Unleashing the Power of Multi-Modal Large Language Models for Spatial Proximity Analysis}



\icmlsetsymbol{equal}{*}

\begin{icmlauthorlist}
\icmlauthor{Jianing Li}{equal,nju}
\icmlauthor{Xi Nan}{equal,pku}
\icmlauthor{Ming Lu}{intel}
\icmlauthor{Li Du}{nju}
\icmlauthor{Shanghang Zhang}{pku}
\end{icmlauthorlist}

\icmlaffiliation{nju}{School of Electronic Science and Engineering, Nanjing University}
\icmlaffiliation{pku}{School of Electronics Engineering and Computer Science, Peking University}
\icmlaffiliation{intel}{Intel Lab China}

\icmlcorrespondingauthor{Shanghang Zhang}{}

\icmlkeywords{Machine Learning, ICML}

\vskip 0.3in
]



\printAffiliationsAndNotice{\icmlEqualContribution} 

\begin{abstract}
Multi-modal large language models (MLLMs) have demonstrated remarkable vision-language capabilities, primarily due to the exceptional in-context understanding and multi-task learning strengths of large language models (LLMs). The advent of visual instruction tuning has further enhanced MLLMs' performance in vision-language understanding. However, while existing MLLMs adeptly recognize \textit{what} objects are in an image, they still face challenges in effectively discerning \textit{where} these objects are, particularly along the distance (scene depth) axis. To overcome this limitation in MLLMs, we introduce Proximity Question Answering (Proximity QA), a novel framework designed to enable MLLMs to analyse the proximity relationship between objects in images. The framework operates in two phases: the first phase focuses on guiding the models to understand the relative depth of objects, and the second phase further encourages the models to analyse the proximity relationships between objects based on their depth perceptions. We also propose a VQA dataset called Proximity-110K, containing additional instructions that incorporate depth information and the proximity relationships of objects. We have conducted extensive experiments to validate Proximity QA's superior ability in depth perception and proximity analysis, outperforming other state-of-the-art MLLMs. Code and dataset will be released at \textcolor{magenta}{https://github.com/NorthSummer/ProximityQA.git}.
\end{abstract} 

\section{Introduction}

In recent years, large language models (LLMs) have catalyzed significant breakthroughs in zero-shot performance across multiple natural language processing tasks. This success is primarily attributed to their exceptional in-context learning and instruction-following capabilities. Building on the advancements in LLMs, multi-modal large language models (MLLMs) have attracted considerable research interest. Typically, an MLLM employs an LLM as its core, responsible for processing and interpreting multi-modal inputs (primarily in vision-language formats) and performing language-based reasoning, aligning more closely with human-like perception towards the world. Recent studies, such as Open-Flamingo \cite{awadalla2023openflamingo}, MiniGPT4 \cite{zhu2023minigpt}, LLaMA-Adapter \cite{zhang2023llama}, Instruct-BLIP \cite{dai2305instructblip}, and LLaVA \cite{liu2023visual}, have demonstrated remarkable capabilities in generating language responses from multi-modal inputs. These advances have elevated performance in various vision-language tasks, including Image Captioning, OCR-Recognition, and Visual Question Answering. However, we analyse that the multi-modal instructions in existing methods predominantly focus on vision-language semantics, almost neglecting or inadequately addressing the geometric aspects. This imbalance enhances models' ability to identify concepts like \textit{What} is in this image but results in a weaker understanding of geometric properties, such as proximity information. Moreover, the development of multi-modal instructions that effectively integrate both semantic and geometric information poses a significant challenge.
%
\begin{figure*}[t!]
\includegraphics[width=17.6cm]{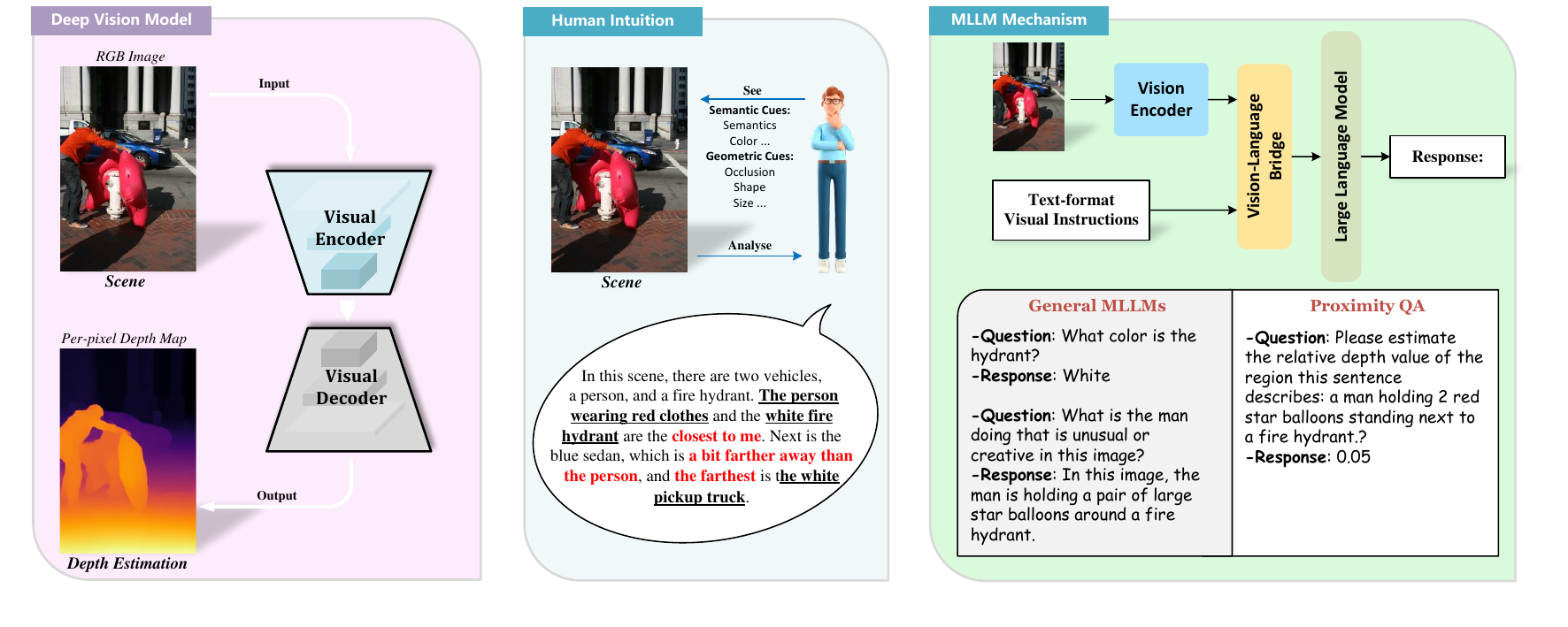}
\centering
\caption{Deep vision models can derive dense geometric information of a scene by estimating accurate depth maps, but humans often understand scenes with both semantic and geometric information. We enable MLLMs to achieve this integrated understanding of semantic and geometric information through multi-modal instructions, thus creating a perception pattern that more closely aligns with human intuition.}
\label{fig:1}
\end{figure*}

Research focused on deriving dense geometric information from a single image using deep models has been ongoing for nearly a decade, initiated by the pioneering work \cite{eigen2014depth}. This problem is defined as the task of estimating per-pixel depth values for an input image, known as Monocular Depth Estimation (MDE). Contemporary state-of-the-art MDE models excel in predicting precise depth maps in various settings, including both outdoor and indoor environments. They have also proven to be highly effective in robust depth estimation across diverse scenes, demonstrating impressive zero/few-shot capabilities. These MDE models are proficient in extracting geometric information from images, effectively addressing the \textit{Where} are they challenge. However, as previously discussed, a comprehensive multi-modal understanding of images requires the integration of both semantic and geometric information, just like human intuition.

To address these challenges, we introduce Proximity-QA, an innovative framework designed to enhance the ability of MLLMs to comprehend geometric information of objects in images via a Question-Answering instruction format. Proximity QA contains two steps to finish training, encompassing perception and reasoning. In the perception step, each object in the image is assigned a relative depth value ranging between 0 and 1. Concurrently, MLLMs are trained to assimilate depth information of these objects through QA-Type instructions. The questions present the semantic information of the objects, while the answers provide their geometric details. By following to these instructions, MLLMs are conditioned to estimate a depth value for the objects. The reasoning step aims to enable the model to infer the proximity relationships between objects within the same image, leveraging its acquired object-level depth estimation skills. For this purpose, a simple yet effective chain-of-thought methodology is incorporated to enhance the model's accuracy in analysing proximity relationships. In summary, Proximity-QA equips MLLMs with the capability to estimate object-level depth information and infer proximity relationships among objects, thereby completing the model's geometric understanding.

Our proposed Proximity QA framework exhibits several significant advantages: a) Geometric Understanding Ability: Proximity-QA effectively addresses the limitations of MLLMs in image geometric perception. By leveraging the instruction-following and reasoning capacities of large language models, our framework is adept at making precise determinations about the proximity relationships of objects within an image.
b) Human-like Perception Pattern: Proximity-QA is uniquely capable of concurrently perceiving both the semantic and geometric information of objects, and articulating this understanding in a human-like manner. We illustrate this logic in Figure \ref{fig:1}. The contribution of this work can be listed as follows:

\begin{itemize}
    \item We propose an unified \textbf{Perception-Reasoning} framework based on MLLMs , namely Proximity QA, for analysing the proximity relationships of objects within an image.
    \item We have collected a dataset for inferring object proximity relationships: Proximity-110K. This dataset comprises two types of VQA conversations, specifically aimed at perceiving object depth and inferring object proximity relationships.
    \item We conducted comparisons with the state-of-the-art MLLMs, demonstrating that our framework possesses unique advantages in inferring the proximity of objects.
\end{itemize}




\section{Related Work}
\subsection{Multimodal Large Language Models}

\begin{figure*}[ht]
\includegraphics[width=17.3cm]{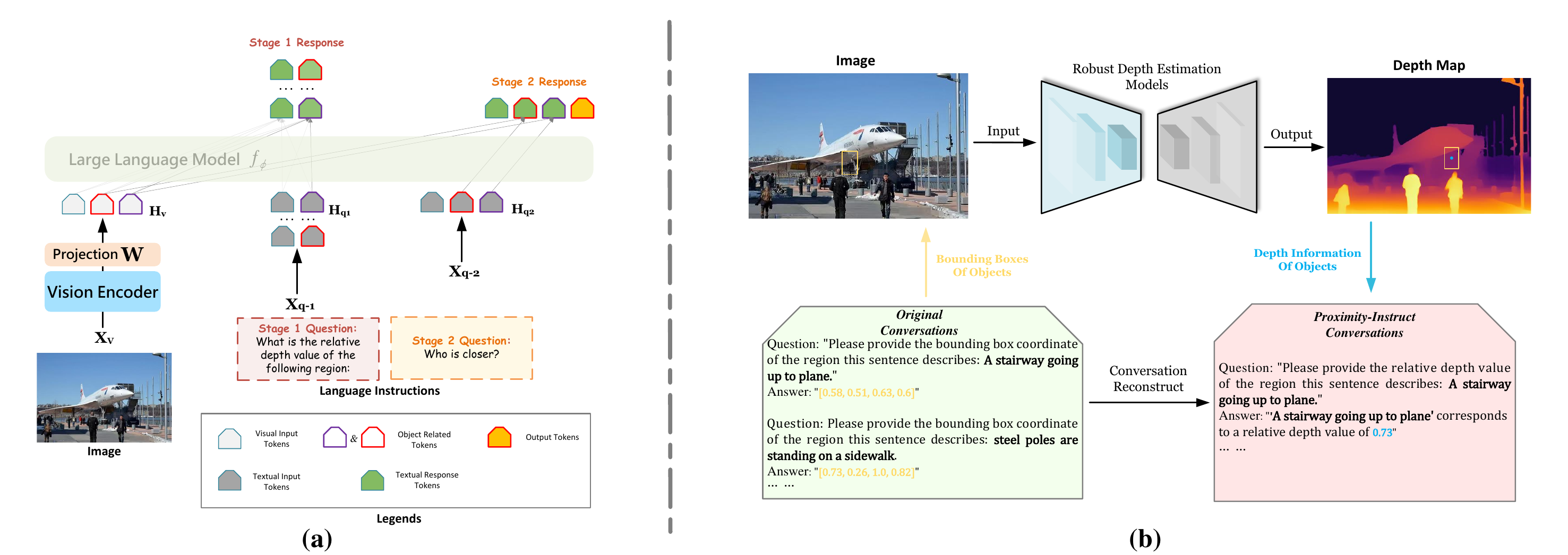}
\centering
\caption{Network architecture of Proximity QA in \textbf{part (a)} and the construction pipeline of Proximity-110K in \textbf{part (b)}.  We adopted a two-stage visual instruction tuning approach to achieve proximity relationship analysis of objects in the image. In the generation of Proximity-110K, we incorporate depth information into the original conversations and build new instructions.}
\label{fig:2}
\end{figure*}

The advancements in computer vision and natural language processing spark the emergence of multimodal large language models (MLLMs) that integrate visual and linguistic capabilities for improved cross-modality understanding. As a pioneer attempt, CLIP\cite{radford2021learning} broadened the scope of language models to include vision-language tasks. The focus has been increasingly on leveraging the strengths of Large Language Models (LLMs). Notably, Flamingo\cite{awadalla2023openflamingo} extensive image-text pairs for cross-modality alignment, enhancing learning effectiveness. BLIP2\cite{li2023blip} introduces a Query-Transformer (Q-Former), which extracts features from a frozen vision encoder, acting as a bottleneck between the vision encoder and the LLM. To further capitalize on these pre-trained models, InstructBLIP\cite{dai2305instructblip} and MiniGPT-4\cite{zhu2023minigpt} create high-quality multi-modal instruction pairs based on BLIP-2, achieving superior performance. Simultaneously, LLaVA\cite{liu2023visual} applies a simple linear projector with minimal learnable parameters for aligning the image and text domains, demonstrating strong performance with specialized instruction data. LLaMA-Adapter\cite{zhang2023llama} inserts a Zero-initialized Attention layer into LLaMA\cite{touvron2023llama}, facilitating multi-modal instruction tuning abilities for a 7B LLaMA. It is noteworthy that recent emerging works have begun to employ LLMs in some traditional visual tasks, such as semantic segmentation\cite{lai2023lisa}, object detection\cite{zang2023contextual}\cite{wang2023visionllm}, and visual groundingc\cite{wang2023visionllm}\cite{lin2023sphinx}, etc. These works demonstrate that LLMs are able to assist visual models in achieving improved zero-shot and open-vocabulary perception performance.

\subsection{Visual Question Answering}
First introduced in\cite{antol2015vqa}, Visual Question Answering (VQA) is a multi-modal task that requires the integration of computer vision and natural language processing techniques. The problem of VQA can be describe as: given an image and a question related to the image, a vision-language model is required to understand the textual question and analyze the content of the image to provide the correct answer. 
Traditional vision-language models typically employ a CNN as the vision encoder and utilize RNN-based\cite{biten2019scene}, GRU-based\cite{cadene2019murel}, or LSTM-based\cite{ben2017mutan}\cite{yu2019deep} models as the language encoder. Finally, they apply specific fusion strategies to integrate vision and language features for generating the final response. Numerous datasets for VQA have been proposed. Categorized by the theme of the question, there are commonsense VQA datasets\cite{zhang2016yin}\cite{zhu2016visual7w}\cite{krishna2017visual}, spatial relationship VQA datasets\cite{johnson2017clevr}, and scientific VQA dataset\cite{lu2022learn}, etc.


\section{Proximity Question and Answering}

\subsection{Problem Defination}

Generally, a MLLM $F$ take as input an image with dimensions $3 \times H \times W$ and a text sequence $T^{In}$ , generating a textual response $T^{Out}$ as the output. This is formalized as:

\begin{equation}
    T^{Out} = F(I, T^{In})
    \label{eq:1}
\end{equation}

Leveraging MLLMs to analyse the spatial proximity relationships of objects in images is a crucial and challenging problem. Existing methods have partially overlooked this aspect in training the MLLMs, highlighting the necessity for more effective and accurate strategies to realise a comprehensive image understanding. Our objective is to empower MLLMs to perceive the relative distances, or depth values between objects in images, thereby enabling the model to more accurately infer the proximity relationship of objects in the image. In other words, we aim to guide the MLLMs to \textbf{speak out} the answer of the question: \textit{How close is it? (Where is it?)} and \textit{Which is closer?}. To define this problem more precisely, we quantify the objective as follows:

Given N ($N \geq 2$) objects $\mathbf{O} = \{ O_1, O_2, ..., O_N \} \in I$ within the image $I$, and a question $Q$ about the proximity relationship between two selected objects $\{ O_s, O_t \} \subseteq \mathbf{O}$, the model generates a corresponding answer $A$ in response to the multi-modal inputs. Deriving from Eq. \ref{eq:1}, this process can be formulated as:

\begin{equation}
     A = F(I, Q(O_s, O_t)) 
\end{equation}

To achieve this objective, we introduce a two-stage QA framework designed to guide MLLMs in analysing the proximity relationship between objects. The first stage involves asking the model to estimate a relative depth values ranging between 0 and 1 of specific objects in the image. Subsequently, in the second stage, we select two objects within a image and instruct the model to analyse their proximity relationship based on their estimated depth values. The following subsection provides a detailed description of this process.




\subsection{Framework Architecture and Training Scheme}
Our framework is basically built upon LLaVA. More specifically, a LLM  is employed to process instructions from textual inputs. Concurrently, a Vision Transformer model pre-trained by CLIP is chosen as the vision encoder. The visual tokens are passed through a 2-layer Multi-Layer Perceptron (MLP) and transformed into the language space, aligning with the textual instruction tokens. Subsequently, these tokens are collectively fed into the LLM to generate responses. We provide an illustration of our framework in the \textbf{(a)} part of Figure \ref{fig:2}.

 
Traditional VQA frameworks tend to directly use the connection between questions and answers to instill the in-context knowledge to the model. However, we argue that this approach is sub-optimal for addressing proximity-related problems. This is because geometric information often lies hidden within or behind the image content, making it challenging to be directly captioned. Hence, directly using Q-A-format instructions to train a MLLM for analysing proximity may fails to achieve the expected performance. To effectively develop such a model, we propose the following two-stage training scheme:


\begin{table}[htbp]
\centering
\begin{tcolorbox}[
    enhanced,
    arc=3mm, 
    outer arc=3mm,
    boxrule=0.8pt, 
    colback=gray!20, 
    colframe=black, 
    coltext=black, 
    boxsep=5pt, 
    left=5pt,
    right=5pt,
    top=5pt,
    bottom=5pt
]
\textbf{X\textsubscript{system-message} }\\
Question:  $Q\textsubscript{1}^{\text{stage1}}$ \text{\textcolor{blue}{(What's the relative depth value of $\mathbf{O_1}$) }}  \\
Answer: $A_1^{\text{stage1}}$ \text{\textcolor{blue}{($\mathbf{D_1}$)}}

\end{tcolorbox}
\captionof{table}{A template for depth perception instructions in Proximity-110K dataset. $Q\textsubscript{1}^{\text{stage1}}$ denotes the 1st question of a scene for the perception stage, while $A_1^{\text{stage1}}$ denotes the answer of $Q\textsubscript{1}^{\text{stage1}}$. The \textbf{X\textsubscript{system-message}} is set for LLMs to better understand the task. }
\label{tab:1} 
\end{table}

\textbf{Stage 1: Perception} In the first stage of our framework, we focus on enabling the model to estimate the distances of objects within an image, guided by specific instructions. For training purposes, we employ straightforward conversation templates. In these templates, the questions inquire about the relative depth value of objects $\mathbf{O}$ in the image, with the answers being a two-digit floating-point number $\mathbf{D}$, normalized between 0 and 1, to represent the depth label of the object. Taking advantage of the instruction-following capabilities of the LLM in our framework, we guide the model to generate the expected relative depth values for objects. In terms of object depth labeling, we utilize MiDAS\cite{ranftl2020towards} to estimate scene disparity. This disparity is then inverted to depth space and get normalized. This stage is dedicated to guiding the model in recognizing objects from images and estimating their depth information, hence we call this stage as the perception stage. Table \ref{tab:1} illustrates an example of the conversation template for the perception stage.


\begin{table}[htbp]
\centering
\begin{tcolorbox}[
    enhanced,
    arc=3mm, 
    outer arc=3mm,
    boxrule=0.8pt, 
    colback=gray!20, 
    colframe=black, 
    coltext=black, 
    boxsep=5pt, 
    left=5pt,
    right=5pt,
    top=5pt,
    bottom=5pt
]
\textbf{X\textsubscript{system-message} }\\
Question: $Q\textsubscript{1}^{\text{stage2}}$ (\textcolor{blue}{Which object seems more approachable? `$\mathbf{O_1}$' or `$\mathbf{O_2}$'}.) \\
Answer: $A_1^{\text{stage2}}$ (\textcolor{blue}{`$\mathbf{O_1}$' corresponds to a relative depth value of $\mathbf{D_1}$, and `$\mathbf{O_2}$' corresponds to a relative depth value of $\mathbf{D_2}$. Since $\mathbf{D_1 > D_2}$, it can be inferred that the object: `$\mathbf{O_2}$' is closer.} )

\end{tcolorbox}
\captionof{table}{A template for proximity analysis instructions is included in the Proximity-110K dataset, where the reasoning process is built upon the depth perception results enhanced during the first stage.}
\label{tab:2} 
\end{table}

\textbf{Stage 2: Reasoning} The second stage is dedicated to enabling the model to infer the proximity relationships between objects, based on their depth perception results for objects within the image. Building upon the perceptual outputs of the first stage, these results provide a solid foundation for further analysis of proximity relationships between objects. Utilizing the instruction-following and reasoning abilities of the LLM in our model, we integrate a straightforward chain-of-thought into the answers of the conversation. This method encourages the model to analyse the proximity relationships between objects, taking into account their estimated depth information. Specifically, we consider two objects in the image, denoted as $O_s$ and $O_t$, along with their respective depth perception values, $D_s$ and $D_t$. The relative proximity is determined by comparing $D_s$ and $D_t$. This comparison results in one of three relational scenarios: $\mathbf{O_s}$ \textbf{being closer}, $\mathbf{O_t}$ \textbf{being closer}, or \textbf{Equally close}. These scenarios correspond to the conditions $D_s < D_t$, $D_s > D_t$, and $D_s = D_t$, respectively. Table \ref{tab:2} presents a conversation template for the reasoning stage, with $s=1$ and $t=2$.



\section{Dataset: Proximity-110K}

\subsection{Data Source} 

Like other VQA datasets, Proximity-110K consists of a collection of images paired with corresponding conversations. The images for this dataset are carefully selected from the Visual Genome \cite{krishna2017visual} and COCO \cite{lin2014microsoft} datasets. However, constructing QA-type conversations, particularly with depth information and object proximity relationships, presents a significant challenge. This complexity stems from the necessity to accurately parse and interpret the depth and proximity information contained within a wide array of images. To address this, we employed an off-the-shelf approach to estimate the depth of objects in the images. Using the robust capabilities of MiDAS\cite{ranftl2020towards}, we estimate the depth maps for the images. Then, we focused on the central points of objects identified by bounding boxes annotations, assuming that the depth value at these central coordinates reflects the object's overall distance. This approach allowed us to integrate depth information into our dataset, facilitating the generation of conversations that accurately reflect the proximity relationships between objects in an image. We selected a total of 110,261 images from the COCO and VG datasets, each annotated with bounding boxes of objects, facilitating the integration of depth information.

\subsection{Conversation Generation}
\textbf{Question Generation} To facilitate the generation of questions for conversations within our dataset, we employed artificially designed question templates. For the perception stage three distinct templates are introduced to direct the model towards estimating the depth values of objects in the images. For example, one such template is: \texttt{"What is the relative depth value of the following region: $<$object caption$>$?"}, where \texttt{$<$object caption$>$} refers to a descriptive sentence or phrase captioning the object. In the reasoning stage, we crafted  twenty question templates. These templates are designed to guide the model to infer or analyse the spatial proximity relationships between objects based on the depth information perceived in the first stage. For instance, a typical question in this stage might be: \texttt{"In this image, which is closer to me, $<$Object1 caption$>$ or $<$Object2 caption$>$?"}. This format aims to enable the model to integrate its perceptual results with spatial reasoning capabilities.


\noindent \textbf{Answer Generation} After preparing the questions, we proceeded to construct the corresponding answers. In the perception answers, where the focus is on estimating depth values, any real number between (0, 1) could be a potential answer. To streamline the estimation process, we limit the relative depth values to two decimal places, which are then utilized as answers to the questions for this stage. For the reasoning answers, we adopt a similar approach by employing standardized sentence templates to structure the model's reasoning process. This method aids in enhancing the consistency of the model's outputs, thereby enabling the model to generate more authentic and effective responses. Part \textbf{(b)} of Figure \ref{fig:3} provides a pipeline of generating conversations for Proximity-110K.

\subsection{Statistics and Analysis}

\textbf{Statistics} In Proximity-110K, there is a total of 559,952 Q-A pairs related to object depth information and 429,925 Q-A pairs focusing on object proximity relationships. On average, the questions pertaining to object depth information contain approximately 16.65 words, while those concerning object proximity relationships average around 14.38 words. Notably, due to the inclusion of the reasoning process in the responses to proximity relationship queries, the average length of these answers extends to 43.08 words.

\begin{figure}[h]
\includegraphics[width=8.2cm]{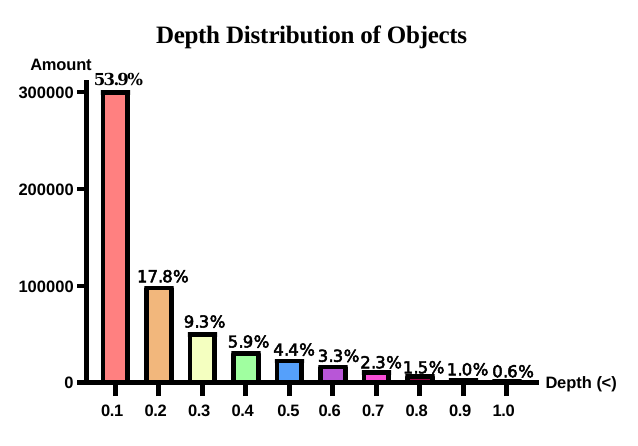}
\centering
\caption{We calculate the distribution of object amounts by depth in the Proximity-110K dataset illustrated in the histogram. The horizontal axis of the histogram denotes depth intervals, while the vertical axis indicates the amount of objects.}
\label{fig:3}
\end{figure}

\noindent \textbf{Distribution} We analyzed the content distribution of questions and answers in the Proximity-110k dataset. Among the perception answers, the proportion of answers with relative depths between 0 and 0.1 was the highest, accounting for 53.89\%; whereas answers with relative depths between 0.9 and 1 constituted the smallest proportion, at 0.61\%. We represented this distribution in a histogram, which reveals that the depth distribution of objects exhibits a long-tail distribution, indicating a predominant amount of objects located closer in distance. Regarding the reasoning answers, answers indicating $O_1$ is closer than $O_2$ accounted for 40.62\%, while responses stating $O_1$ is farther than $O_2$ comprised 40.43\%. Answers suggesting that $O_1$ and $O_2$ are at the equal proximity constituted 18\%.  In this context, $O_1$ and $O_2$ refer to the first and second objects mentioned in a sentence, respectively, as can be referred in Figure \ref{fig:3}.

\noindent \textbf{Correctness} To assess the correctness of Proximity-110K,  we manually selected 25 images-text pairs to evaluate the quality of the corresponding conversations. To sum up, approximately 7.5\% of the conversations in the dataset have the potential for more accurate answers. Our assessment focuses on the following aspects: 

\begin{itemize}
    \item Offset of the Bounding Box Center Points: We use the center points of bounding boxes as coordinates to locate objects and obtain their depth information. However, in specific scenarios, these center points might deviate from the actual surface of the object due to occlusions or the presence of an excessively large background area.

    \item Single Annotation for Multiple Objects: In some cases, images contain multiple objects of the same category (semantic class) with high feature overlap. For these instances, textual annotations should provide clear captions to differentiate each object. If multiple objects are annotated without sufficient distinction, it could result in inaccuracies or hallucinations after training the MLLMs.
\end{itemize}

\begin{table*}[p]
\centering
\begin{minipage}{\textwidth}
\centering
\caption{Qualitative comparison of depth perception and proximity analysis capabilities with the state-of-the-art MLLMs.}
\begin{tabularx}{\textwidth}{X|X}
\toprule
Example Chatting 1 & Example Chatting 2 \\
\midrule
\\
\hspace{.05em} \quad  \includegraphics[width=0.42\textwidth]{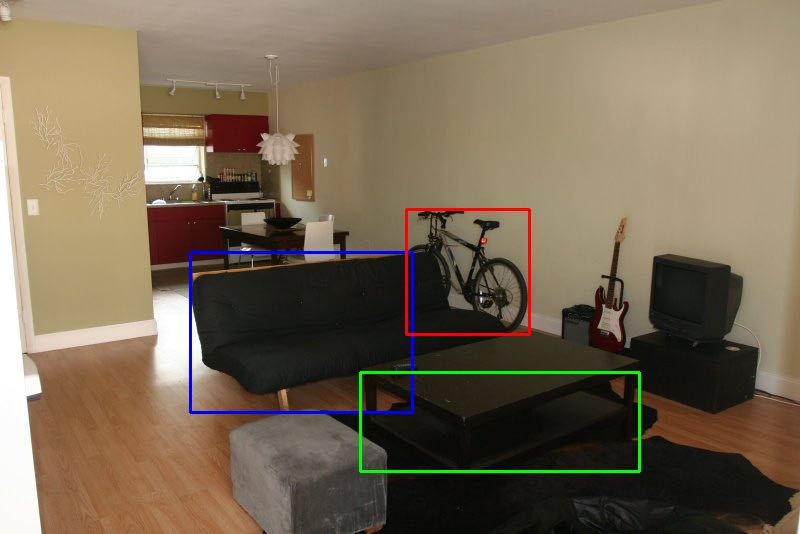} & \hspace{1em} \includegraphics[width=0.42\textwidth]{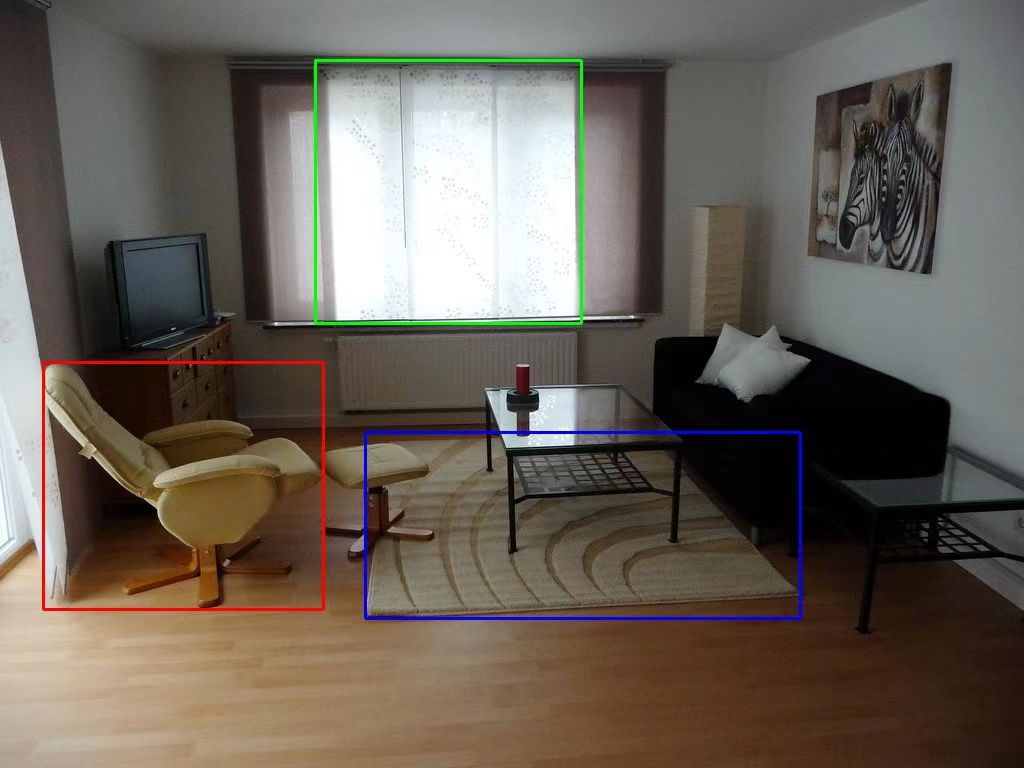} \\
\\

\midrule
\large \textbf{User}: What is the relative depth value of frame in the image?. & \large \textbf{User}: What is the relative depth value of the following region: rug. \\
\textbf{Proximity QA}: 0.29  &  \textbf{Proximity QA}:  0.35 \\
\\
\large \textbf{User}: Please estimate the depth of the frame within the scene, outputting this estimate as a value ranging from 0 to 1, where 0 represents the closest point to the viewer and 1 represents the farthest point. &  \large \textbf{User}:Please estimate the depth of the rug within the scene, outputting this estimate as a value ranging from 0 to 1, where 0 represents the closest point to the viewer and 1 represents the farthest point. \\
 \textbf{LLaVA-1.5-7B}: 0.5 & \textbf{LLaVA-1.5-7B}: 0.5 \\
 \textbf{Instruct-BLIP-6.7B}: 0 &  \textbf{Instruct-BLIP-6.7B}: 0.5\\
\textbf{Qwen-VL-7B}: 0 &  \textbf{Qwen-VL-7B}: 0.7\\
\\
\textcolor{green}{Ground Truth} : 0.16 &  \textcolor{green}{Ground Truth} : 0.21 \\
\midrule

\large \textbf{User}: Which is closer, 'shelf' or 'bicycle'? Answer the question using depth perception and proximity reasoning. & \large \textbf{User}: Which is closer, 'curtains' or 'chair'? Answer the question using a single word or phrase. \\
\textbf{Proximity QA}:  'shelf' corresponds to a depth of 0.04, and 'bicycle' corresponds to a depth of 0.45. since 0.45 $>$ 0.04, it can be inferred that the object: 'shelf' is closer, the answer is: 'shelf'. \textcolor{green}{\ding{51}} &  \textbf{Proximity QA}: chair \textcolor{green}{\ding{51}} \\
\\
\large \textbf{User}: Which is closer, 'shelf' or 'bicycle'? Answer the question using a single word or phrase. &  \large \textbf{User}: Which is closer, 'curtains' or 'chair'? Answer the question using a single word or phrase. \\
\textbf{LLaVA-1.5-7B}: bicycle \textcolor{purple}{\ding{55}} &  \textbf{LLaVA-1.5-7B}: chair \textcolor{green}{\ding{51}} \\
\textbf{Instruct-BLIP-6.7B}: bicycle \textcolor{purple}{\ding{55}} & \textbf{Instruct-BLIP-6.7B}: curtains \textcolor{purple}{\ding{55}} \\
\textbf{Qwen-VL-7B}: bicycle \textcolor{purple}{\ding{55}} & \textbf{Qwen-VL-7B}: chair \textcolor{green}{\ding{51}} \\
\\
\textcolor{green}{Ground Truth} : shelf & \textcolor{green}{Ground Truth} : chair \\
\bottomrule
\end{tabularx}
\label{tab:3}
\end{minipage}
\end{table*}

\begin{table*}[t]
\centering
\centering
\caption{Comparison of the perception performance on the GQA-Conversion validation set with state-of-the-art MLLMs.}
\setlength{\tabcolsep}{2.05mm}
\scalebox{1.0}{
\begin{tabular}{>{\centering\arraybackslash}p{2.5cm}|>{\centering\arraybackslash}p{2.5cm}|>{\centering\arraybackslash}p{2.5cm}ccccccc}
\hline\hline
\multirow{2}{*}{Method} & \multirow{2}{*}{ LLM Size} & \multicolumn{7}{c}{$GQA$ Results} \\
 &  & Valid A. Ratio ↑ & MSE ↓ & RMSE ↓ & Sq Rel ↓ & $\delta 1 $ ↑ & $\delta 2 $ ↑ & $\delta 3 $ ↑ \\
\hline
\hline
\multirow{2}{*}{LLaVA-1.5} & Vicuna-7B & \textbf{99.98\%} & 0.139  & 0.373 &  4.189 & 0.083 & 0.164 & 0.238  \\
 & Vicuna-13B &  77.13 \%  &  0.139 & 0.372 &  4.169 & 0.088 & 0.170 & 0.248 \\
\hline
\multirow{2}{*}{BLIP2} & OPT-6.7B &  98.06 \% & 0.122 & 0.349 & 3.125 & 0.050 & 0.094 & 0.137 \\
 & OPT-2.7B &  98.06 \% & 0.122 & 0.349 & 3.125 & 0.050 & 0.094 & 0.137 \\
\hline
InstructBLIP & Vicuna-7B  &  96.42 \% & 0.116 & 0.340 & 2.854 & 0.043 & 0.077 & 0.108 \\
\hline
QWen-VL & Qwen-7B &  99.55 \% & 0.107 & 0.322 & 1.443 & 0.008 & 0.016 & 0.025 \\
\hline
Proximity QA  & Vicuna-7B  & 91.75 \% & \textbf{0.022} & \textbf{0.147} & \textbf{0.231} & \textbf{0.256} & \textbf{0.475} & \textbf{0.609} \\
\hline\hline
\end{tabular}}
\label{tab:4}
\end{table*}

\section{Experiments}
\subsection{Settings}

\textbf{Implementation Details} In terms of model selection, we have aligned with LLaVA-1.5, utilizing Vit-L/336 as the visual backbone of the model, employing Vicuna-7B\cite{chiang2023vicuna} as the LLM within the model, and implementing a 2-layer MLP as the projector for aligning the visual modality with the linguistic modality. Our model was trained using 8 * Tesla V100 GPUs. Initially, the model underwent pre-training on the CC-595K\cite{sharma2018conceptual} dataset for one epoch to obtain a projector for modality alignment. Subsequently, we fine-tuned the model using LoRA\cite{hu2021lora} on both LLaVA-665K\cite{liu2023improved} and Proximity110K for one epoch, with a learning rate of 2e-5 and a batch size of 12.

\noindent \textbf{Evaluation Data} Given the current absence of any benchmarks or datasets related to proximity VQA for MLLMS, we have opted to convert publicly available benchmark datasets for evaluation. GQA\cite{hudson2019gqa} is a high-quality VQA dataset designed from real-world scenarios, providing object annotations at the bounding-box level. We filter out  Q-A pairs that contain object bounding box annotations from the GQA validation set. Using these bounding boxes, we constructed new proximity-related Q-A pairs, following a methodology which is used in the construction of our Proximity-110K dataset. In total, there are 9912 perception Q-A pairs and 8410 proximity Q-A pairs in the converted GQA dataset. Additionally, we selected 39 images from the Make3D\cite{saxena2008make3d} dataset to construct another evaluation dataset containing 39 Q-A pairs. Make3D provides depth Ground-Truth captured by depth cameras, which significantly enhances the correctness of the constructed proximity relationships, thereby yielding more reliable assessment results.

\subsection{Qualitative Results}
In Table \ref{tab:3}, we showcase the qualitative results of Proximity QA in comparison with other MLLMs in answering the questions about object proximity relationships. We select two images from GQA dataset, and visualize their experimental results. In each visualization, we query two types of questions to the MLLMs, asking the models to answer about the \textbf{relative depth value} of objects and the \textbf{proximity relationship} between the objects in the image. We selected InstructBLIP\cite{dai2305instructblip}, LLaVA\cite{liu2023visual}, and Qwen-VL\cite{bai2023qwen} as baselines. It is important to note that, due to the weaker depth perception capabilities of these baselines, we employed more detailed questions to prompt them for more structured responses. This approach aimed to overcome the limitations in their depth perception abilities, thereby enabling a more effective comparison of their capabilities with those of Proximity QA.

\begin{table}[h]
	\centering
	\begin{center}
	\caption{Comparison to the state-of-the-art MLLMs on inferring proximity relationships on the GQA-Conversion validation set.}
	\setlength{\tabcolsep}{2.05mm}{
    \scalebox{0.90}{
	\begin{tabular}{c|c|cc}
				\hline\hline
    
                \multirow{2}{*}{Method} & \multirow{2}{*}{LLM Size} & \multicolumn{2}{c}{$GQA$ Results} \\
                 &  & Valid A. Ratio ↑ & Accuracy ↑  \\    
				\hline
				\hline
				BLIP2 & OPT-6.7B &  99.83 \% & 43.20 \% \\
                InstructBLIP & Vicuna-7B  &  98.06 \% & 43.32 \% \\
                \hline
				QWen-VL & Qwen-7B &  99.85 \% & 42.28 \%  \\            
				\hline\hline

				Proximity QA  & Vicuna-7B  & \textbf{99.89 \%}  &   \textbf{43.62 \%} \\ 
             \hline\hline
		\end{tabular}}}
		\label{tab:5}
	\end{center}
\end{table}

These two examples demonstrate that Proximity QA is capable of providing more reliable and standardized responses in terms of object depth perception in the image, thereby more accurately answering the questions. In contrast, the baselines are prompted utilizing detailed questions, however the models fail to generate more reliable perceptual results. Furthermore, in terms of inferring the proximity relationship of objects, Proximity QA can reason out the results through its depth perception ability, exhibiting stronger explainability. 


\begin{table}[h]
	\centering
	\begin{center}
	\caption{Comparison to the state-of-the-art MLLMs on inferring proximity relationships on the Make3D-Conversion validation set.}
	\setlength{\tabcolsep}{2.05mm}{
    \scalebox{0.90}{
	\begin{tabular}{c|c|cc}
				\hline\hline
    
                \multirow{2}{*}{Method} & \multirow{2}{*}{LLM Size} & \multicolumn{2}{c}{$Make3D$ Results} \\
                 &  & Valid A. Ratio ↑ & Accuracy ↑  \\    
				\hline
				\hline
    			LLaVA-1.5 & Vicuna-7B &  74.36 \%  &  48.71 \%  \\
                \hline
				\multirow{2}{*}{BLIP2}  & OPT-2.7B &  76.92 \% & 33.33 \% \\
                                    & OPT-6.7B &  66.66 \% & 25.64 \% \\
                \hline
                InstructBLIP & Vicuna-7B  &  \textbf{79.48} \% & 28.20 \% \\
                \hline
   
				\hline\hline

				Proximity QA  & Vicuna-7B  & \textbf{79.48 \%}  &   \textbf{51.28 \% } \\ 
             \hline\hline
		\end{tabular}}}
		\label{tab:6}
	\end{center}
\end{table}

\subsection{Quantitative Results}

To comprehensively evaluate the performance of Proximity QA and other MLLMs on tasks of depth perception and proximity estimation, we employed various metrics to demonstrate its proficiency in both Perception and Reasoning.

On the depth perception task, we required the model to output a numerical value representing the estimated relative depth of an object. However, during the visualization of experimental results, we observed that even with detailed prompts, MLLMS were not always able to produce a standardized response, namely a decimal between 0 and 1. Consequently, we measured two key metrics. The first is \textbf{Valid Answers Ratio} (Valid A. Ratio in Table \ref{tab:4}-\ref{tab:6}), which quantifies the proportion of standardized responses provided by the model out of all responses. The second metric is \textbf{Perception Errors}, for which we utilized assessment criteria from the visual MDE task\cite{eigen2014depth} to quantify the depth perception performance of MLLMS for all valid estimations. It is noteworthy that our evaluation focuses only on the perception errors of specific objects, rather than calculating the perception errors of the entire scene. In Table \ref{tab:4}, we present a comparative analysis of these metrics against baselines on GQA-Conversion dataset. Proximity QA achieved superior results across all metrics on Perception Errors compared to the baseline models. The performance of Proximity QA is particularly notable in MSE (0.022) and Sq Rel (0.231), reflecting its improved capability in depth perception.

In terms of proximity analysis, we also conducted comparisons with state-of-the-art MLLMS. The metrics used for this comparison included  \textbf{Valid Answers Ratio} and \textbf{Accuracy}. Accuracy represents the ratio of correct answers among all generated responses of inferring the proximity relationships. We conducted evaluations on the GQA-Conversion and Make3D-Conversion datasets, with the results presented in Table \ref{tab:5} and Table \ref{tab:6}. On the GQA-Conversion dataset, we achieve a valid answers ratio of 99.89\% and an accuracy of 43.62\%. On the Make3D-Conversion dataset, we obtain a Valid Answers Ratio of 79.48\% and an Accuracy of 51.28\%, demonstrating remarkable generalization capabilities of Proximity QA in proximity analysis.

\section{Conclusion}
In this work, we present Proximity Question Answering (Proximity QA), a novel framework that effectively enhances the spatial depth perception and proximity analysis capabilities of multi-modal large language models (MLLMs), addressing a critical limitation in current MLLMs. Proximity QA enables the MLLM to accurately analyse the proximity relationship between objects in images, thereby accomplishing an integrated understanding of scene semantics and geometry. This is achieved through a two-stage visual-instruction-tuning process: the first stage focuses on perceiving the relative depth of objects, while the second stage leverages this perception ability to reason out the object proximity relationships. Additionally, we propose a Visual Question Answering (VQA) dataset, Proximity-110K, to support relevant research. Our comprehensive experiments on two converted datasets demonstrate Proximity QA's superiority over existing state-of-the-art MLLMs in perceiving depth information and conducting proximity analysis, marking a significant advancement in the geometric understanding of MLLMs.


\bibliography{references}
\bibliographystyle{icml2023}

\appendix
\onecolumn
\section{Additional Related Work}
\subsection{Monocular Depth Estimation}

In Monocular Depth Estimation (MDE), 'Depth' denotes the distance from an object's surface within an image to the observer (or camera) capturing the scene. The seminal work of Eigen et al. \cite{eigen2014depth} is among the initial efforts that spurred recent advancements in MDE. They introduced a novel two-stage architecture, coarse and fine, treating depth estimation as a pixel-level regression problem. Similar to semantic segmentation tasks, a prevalent approach in MDE is employing an encoder-decoder structure, which incorporates CNNs \cite{Xu_2018_CVPR}\cite{Ramamonjisoa_2020_CVPR}\cite{lee2019monocular}\cite{Ramamonjisoa_2019_ICCV}\cite{fu2018deep}\cite{godard2017unsupervised} or transformers \cite{ranftl2021vision}\cite{yang2021transformer}. The encoder captures contextual information and learns a global representation, while the decoder strives to establish a connection between context, texture, and depth information. This process is often facilitated by full-supervision or self-supervision \cite{godard2017unsupervised}\cite{guizilini20203d}\cite{lyu2020hr}. Moreover, innovations in regression techniques \cite{fu2018deep}\cite{bhat2021adabins}\cite{roy2016monocular} have enhanced the efficiency of representing depth information. Recent research highlights the significant potential of integrating MDE with auxiliary tasks like semantic segmentation \cite{Jiao_2018_ECCV}\cite{hoyer2021three}.

In our submitted paper, we describe depth information as a crucial type of geometric information. Typically, geometric information of an entire scene is constituted by various elements, including depth, shape, size, and surface normal information. However, we focus on depth information due to its dense format and the extensive development of depth estimation tasks. These characteristics render depth information as the most representative geometric element in images, providing a comprehensive understanding of the scene's spatial structure.

\begin{table*}[t]
\centering
\begin{tcolorbox}[
    enhanced,
    arc=3mm, 
    outer arc=3mm,
    boxrule=0.8pt, 
    colback=gray!20, 
    colframe=black, 
    coltext=black, 
    boxsep=5pt, 
    left=5pt,
    right=5pt,
    top=5pt,
    bottom=5pt
]
\Large\textbf{ Question Templates}\\

\large\textbf{For Depth Perception}\\
$\mathbf{Q_{1-1}}$: What's the relative depth value of region: $R_1$ in the image? \\
$\mathbf{Q_{1-2}}$: Please provide me with the relative depth value of region: $R_1$ in the picture. \\
$\mathbf{Q_{1-3}}$: Please estimate the relative depth value of region: $R_1$ in the image. \\

\large\textbf{For Proximity Analysis}\\
\normalsize
Below templates are for direct answers:\\
\small
$\mathbf{Q_{2-1}}$: Is Region1: $R_1$ nearer to us, or Region2: $R_2$ nearer to us? Answer the question using a single word or phrase. \\
$\mathbf{Q_{2-2}}$: Which region is closer, Region1: $R_1$ or Region2: $R_2$? Answer the question using a single word or phrase. \\
$\mathbf{Q_{2-3}}$: Is Region1: $R_1$ closer, or Region2: $R_2$ closer? Answer the question using a single word or phrase. \\
$\mathbf{Q_{2-4}}$: Please tell me which region is closer to me, Region1: $R_1$ or Region2: $R_2$? Answer the question using a single word or phrase. \\
$\mathbf{Q_{2-5}}$: Please determine which is closer, Region1: $R_1$ or Region2: $R_2$? Answer the question using a single word or phrase. \\
$\mathbf{Q_{2-6}}$: In this image, which is closer to me, Region1: $R_1$ or Region2: $R_2$? Answer the question using a single word or phrase. \\
$\mathbf{Q_{2-7}}$: Which region seems more approachable, Region1: $R_1$ or Region2: $R_2$? Answer the question using a single word or phrase. \\
$\mathbf{Q_{2-8}}$: Which of the two regions is closer, Region1: $R_1$ or Region2: $R_2$? Answer the question using a single word or phrase. \\
$\mathbf{Q_{2-9}}$: In this picture, which region is more approachable, Region1:$R_1$ or Region2: $R_1$? Answer the question using a single word or phrase. \\
\normalsize
Below templates are for reasoning answers:\\
\small
$\mathbf{Q_{2-10}}$: Is Region1: $R_1$ nearer to us, or Region2: $R_2$ nearer to us? Answer the question using depth perception and reasoning.
$\mathbf{Q_{2-11}}$: Which region is closer, Region1: $R_1$ or Region2: $R_2$? Answer the question using depth perception and reasoning.
$\mathbf{Q_{2-12}}$: Is Region1: $R_1$ closer, or Region2: $R_2$ closer? Answer the question using a single word or phrase. \\
$\mathbf{Q_{2-13}}$: Please tell me which region is closer to me, Region1: $R_1$ or Region2: $R_2$? Answer the question using depth perception and reasoning. \\
$\mathbf{Q_{2-14}}$: Please determine which is closer, Region1: $R_1$ or Region2: $R_2$? Answer the question using depth perception and reasoning. \\
$\mathbf{Q_{2-15}}$: In this image, which is closer to me, Region1: $R_1$ or Region2: $R_2$? Answer the question using depth perception and reasoning. \\
$\mathbf{Q_{2-16}}$: Which region seems more approachable, Region1: $R_1$ or Region2: $R_2$? Answer the question using depth perception and reasoning. \\
$\mathbf{Q_{2-17}}$: Which of the two regions is closer, Region1: $R_1$ or Region2: $R_2$? Answer the question using depth perception and reasoning. \\
$\mathbf{Q_{2-18}}$: In this picture, which region is more approachable, Region1:$R_1$ or Region2: $R_1$? Answer the question using depth perception and reasoning. \\

\end{tcolorbox}
\captionof{table}{Templates for ``Region" type questions for depth perception and proximity analysis in the Proximity 110K dataset. For all templates, $R_1$ or $R_2$ denote the captions for a region or object. }
\label{tab:sup1} 
\end{table*}

\subsection{Vision Language Models}

Exploring the interaction between vision and language, Vision Language Models (VLMs) play a pivotal role in advancing artificial intelligence research. They represent a critical domain for multi-modal understanding and emulating complex cognitive patterns similar to human perception. Early works focused on using probabilistic models to retrieve keywords or captions to describe images, laying the groundwork for Image Captioning \cite{feng2010many}\cite{farhadi2010every}\cite{kuznetsova2012collective}. Subsequent efforts shifted towards describing explicit geometric visual information, such as the 2D location of objects in images, through language responses, forming the early basis of Visual Grounding \cite{kazemzadeh2014referitgame}\cite{plummer2015flickr30k}.

With the advent of deep learning models, tasks like Visual Question Answering and Visual Reasoning gained prominence. Furthermore, the trend towards unifying models for extracting vision and language information emerged. Specifically, Convolutional Neural Networks (CNNs) became the prevalent choice for visual encoding in VLMs \cite{tan2019lxmert}, while BERT \cite{devlin2018bert} introduced a two-stage training framework of pretraining and finetuning, becoming widely adopted in subsequent researches \cite{lu2019vilbert}\cite{li2019visualbert}\cite{huang2020pixel}. Recently, the Vision Transformer (ViT) has emerged as a new foundational model in vision, replacing CNNs as the visual encoder in VLMs \cite{li2021align}\cite{wang2021simvlm}.

The introduction of GPT-3 \cite{brown2020language} marked a new era in Large Language Models (LLMs) within Natural Language Processing (NLP). It demonstrated remarkable capabilities across a range of NLP tasks, achieved by scaling up the model parameters and dataset size. Alongside ViT, this led to the development of more unified Vision Language Models, with LLMs being incorporated as comprehensive language models in VLMs, culminating in the creation of Multi-modal Large Language Models (MLLMs).


\section{Additional Details of Proximity-110K}

\subsection{Question Templates}
As outlined in our paper, we employed a series of templates to generate questions. Recognizing that object or region captions can vary from a single word to a full sentence, we developed two template types: ``Region" and ``Object". These correspond to scenarios with brief and complex captions, respectively. To dissect the linguistic elements of captions, we utilized SceneGraphParser \cite{schuster2015generating}. Captions comprising just a subject and an attribute were processed using the `Object' type template. In contrast, more complex captions necessitated the utilization of the ``Region" type template. Table \ref{tab:sup1} in our paper provides a comprehensive list of all the ``Region" type templates implemented in our study. For the "Object" type templates, the only difference from the "Region" type templates lies in the prefix of the captions. The sentence structure remains unchanged, merely substituting `Region1: $R_1$' and `Region2: $R_2$' in the template with `$O_1$' and `$O_2$', respectively. Here, $R_1$, $R_2$, $O_1$, and $O_2$ all refer to the captions of objects. Hence, we have a total of 42 templates to construct all questions.


\section{Additional Details about Evaluation}

\noindent \textbf{Depth Perception Evaluation} In our paper, we apply widely-used Monocular Depth Estimation (MDE) evaluation metrics to assess the depth perception error of MLLMs. We chose Mean Squared Error (MSE), Root Mean Squared Error (RMSE), Squared Relative Error (Sq Rel), and accuracy thresholds ($\delta 1$, $\delta 2$, $\delta 3$) as our specified metrics. The estimated depth value of the $i$-th object in all evaluated images is denoted as $D_i$, and $\hat{D_i}$ represents the ground truth depth value of the corresponding object. The detailed formulations of these metrics are listed as follows:

\begin{itemize}
    \item MSE $ = \frac{1}{N}\sum_{i=1}^{N}||D_i - \hat{D_i}||^{2}$
    \item RMSE $ = \sqrt{\frac{1}{N}\sum_{i=1}^{N}||D_i - \hat{D_i}||^{2}}$
    \item Sq Rel $ =  \frac{1}{N}\sum_{i=1}^{N}  \frac{||D_i - \hat{D_i}||^{2}}{D_i}$
    \item $\delta 1$: $\% \hspace{.25em} \text{of} \hspace{.25em}  D_i \hspace{.255em}  \text{s.t.} \hspace{.25em}  max(\frac{\hat{D_i}}{D_i},\frac{D_i}{\hat{D_i}}) < 1.25 $
    \item $\delta 2$: $\% \hspace{.25em} \text{of} \hspace{.25em}  D_i \hspace{.255em}  \text{s.t.} \hspace{.25em}  max(\frac{\hat{D_i}}{D_i},\frac{D_i}{\hat{D_i}}) < (1.25)^2 $
    \item $\delta 3$: $\% \hspace{.25em} \text{of} \hspace{.25em}  D_i \hspace{.255em}  \text{s.t.} \hspace{.25em}  max(\frac{\hat{D_i}}{D_i},\frac{D_i}{\hat{D_i}}) < (1.25)^3 $    
\end{itemize}

\begin{table*}[p]
\centering
\begin{minipage}{\textwidth}
\centering
\caption{Additional qualitative comparison of depth perception and proximity analysis capabilities with the state-of-the-art MLLMs.}
\begin{tabularx}{\textwidth}{X|X}
\toprule
Additional Qualitative Case 1 & Additional Qualitative Case 2 \\
\midrule
\\
\hspace{.50em} \quad \quad  \includegraphics[width=0.38\textwidth]{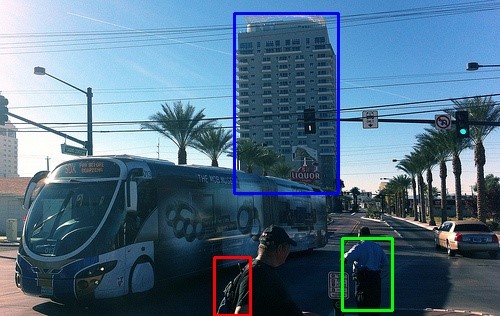} & \quad \hspace{1em} \includegraphics[width=0.38\textwidth]{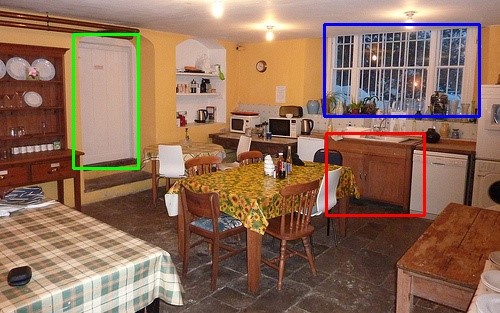} \\
\\
\midrule

\textbf{User}: What is the relative depth value of building in the image?. & \textbf{User}: Please provide me with the relative depth value of window in the picture. \\
\textbf{Proximity QA}: 0.45 & \textbf{Proximity QA}:  0.72 \\
\\
\textbf{User}: Please estimate the depth of a building within the scene, outputting this estimate as a value ranging from 0 to 1, where 0 represents the closest point to the viewer and 1 represents the farthest point. & \textbf{User}: Please estimate the relative depth value of the region this sentence describes: window. \\
\textbf{LLaVA-1.5-7B}: 0.5 & \textbf{LLaVA-1.5-7B}: [0.68, 0.23, 0.99, 0.47] \\
\textbf{Instruct-BLIP-6.7B}: 0 & \textbf{Instruct-BLIP-6.7B}: 10 feet \\
\textbf{Qwen-VL-7B}: 0 & \textbf{Qwen-VL-7B}: (671,108),(941,378) \\
\\
\textcolor{green}{Ground Truth} : 0.64 &  \textcolor{green}{Ground Truth} : 0.6 \\
\midrule

\textbf{User}: Which is closer, 'uniform' or 'backpack'? Answer the question using depth perception and proximity reasoning. & \textbf{User}:  Which is closer, 'door' or 'cabinet'? \\
\textbf{Proximity QA}:  'uniform' corresponds to a depth of 0.04, and 'backpack' corresponds to a depth of 0.01. since 0.01 < 0.04, it can be inferred that the object: 'backpack' is closer, the answer is: 'backpack' \textcolor{green}{\ding{51}} &  \textbf{Proximity QA}: cabinet \textcolor{green}{\ding{51}} \\
\\
\textbf{User}:Which is closer, 'uniform' or 'backpack'? Answer the question using a single word or phrase. &  \textbf{User}: Which is closer, 'door' or 'cabinet'?  \\
\textbf{LLaVA-1.5-7B}: uniform \textcolor{purple}{\ding{55}} &  \textbf{LLaVA-1.5-7B}: the door is closer to the cabinet \textcolor{purple}{\ding{55}} \\
\textbf{Instruct-BLIP-6.7B}: uniform \textcolor{purple}{\ding{55}} & \textbf{Instruct-BLIP-6.7B}: door \textcolor{purple}{\ding{55}} \\
\textbf{Qwen-VL-7B}: backpack \textcolor{green}{\ding{51}} & \textbf{Qwen-VL-7B}: (169,125),(283,535) \textcolor{purple}{\ding{55}} \\
\\
\textcolor{green}{Ground Truth} : backpack & \textcolor{green}{Ground Truth} : cabinet \\
\bottomrule
\end{tabularx}
\label{tab:sup2}
\end{minipage}
\end{table*}

\noindent \textbf{Proximity Analysis Evaluation} We have introduced two key metrics – Valid Answers Ratio and Accuracy – to evaluate the performance of MLLMs in Proximity Analysis Evaluation. For assessing Accuracy, we utilize a precise method of regular expression matching. This involved comparing the answers generated by the model against predefined standard answers for each question. An answer is deemed correct if it matched the standard answer verbatim. We adopted this approach because we expect precise answers, rather than flexible or open-ended responses from the models, as most of the previous VQA frameworks have emphasized. As a result, the preference is for standard and exact answers rather than those based on similarity.

\noindent \textbf{Evaluation Dataset Conversion} In our submitted paper, we construct evaluation datasets by selecting images from the GQA and Make3D datasets. For the GQA dataset, given its inclusion of Q-A pairs with object bounding box information, we adopt the same method to the construction of Proximity-110K for creating Q-A pairs related to perception and proximity analysis. Regarding the Make3D dataset, though it includes Depth GT captured by a depth camera, it lacks object bounding box annotations and corresponding textual captions. To address this, we analyse the images in Make3D using GPT4-V to generate captions for the objects. Subsequently, we acquire the object depth labels using manually annotated central coordinates of the objects and the corresponding image-level depth GT.

\section{Additional Qualitative Results}

We additionally provided two cases of visualizations on the GQA-Conversion dataset. For the first case, considering that the visualizations in the paper were all of indoor scenes, we select an outdoor image to conduct a qualitative performance evaluation. Similar to the prompt method used in the paper, we detail prompts for baseline methods. Specifically, for questions about depth perception, we include the statement, \texttt{"Outputting this estimate as a value ranging from 0 to 1, where 0 represents the closest point to the viewer and 1 represents the farthest point."} For questions about proximity analysis, we add \texttt{"Answer the question using a single word or phrase."} to guide the model in producing standardized responses. Table \ref{tab:sup2} presents the specific responses from Proximity QA and the baselines. It is observable that Proximity QA demonstrates strong capabilities in depth perception and proximity analysis in outdoor scenes as well.

In the second case study, we investigate the impact of prompts on model responses. As previously discussed, we guided the baseline models to generate standardized outputs by introducing detailed prompts. We chose an indoor scene image from the GQA-Conversion Dataset. In this case, we provide the same prompts to both Proximity QA and the baselines, then observed the differences in their outputs. Specifically, for questions about depth perception, we ask all models only about the object's depth, without any additional textual prompts. The results shows that the outputs of the baseline models are in the invalid formats; for instance, LLaVA-1.5-7B responds with [0.68, 0.23, 0.99, 0.47], which appears to be normalized coordinates of a window's bounding box, rather than depth values. Qwen-VL exhibits a similar trend. InstructBLIP output a response of 10 feet, corresponding to a non-standard unit of a absolute depth value, which is also considered an invalid answer. In the aspect of proximity analysis, we also remove the detailed prompt, and the outputs of the baseline models tend to be invalid as well. For example, LLaVA-1.5-7B respond with an incorrect conclusion rather than a specific object, and Qwen-VL once again provide an answer in the form of bounding box coordinates. This phenomenon further corroborates our viewpoint in the paper that existing MLLMs are insufficient in terms of geometry understanding of images. Without detailed explanations for the prompts, the ability of MLLMs to follow instructions is significantly constrained, resulting in numerous invalid responses.

\end{document}